\documentclass[11pt,a4paper]{article}
\usepackage{times,latexsym}
\usepackage{url}
\usepackage[T1]{fontenc}

\usepackage{times}
\usepackage{latexsym}
\usepackage{url}            
\usepackage{booktabs}       
\usepackage{amsfonts}       
\usepackage{nicefrac}       
\usepackage{microtype}      
\usepackage{xcolor}         
\usepackage{latexsym}
\usepackage{amssymb}
\usepackage{amsmath}
\usepackage{booktabs}
\usepackage{enumerate}
\usepackage{graphicx}
\usepackage{subfigure}
\usepackage{xspace}
\usepackage{float}
\usepackage{bbm}
\usepackage{bm}
\usepackage{multirow}
\usepackage{booktabs}
\usepackage{color}
\usepackage{framed}
\usepackage{stfloats}
\usepackage{amssymb}
\usepackage{pifont}
\usepackage{arydshln}
\usepackage{enumitem}
\usepackage{wrapfig}
\usepackage{algorithmic}
\usepackage{algorithm}
\usepackage{makecell}
\usepackage[T1]{fontenc}
\usepackage[utf8]{inputenc}
\usepackage[most]{tcolorbox}
\usepackage{tabularx}
\usepackage{hyperref}
\usepackage{microtype}
\usepackage{inconsolata}
\usepackage{multirow}
\usepackage[T1]{fontenc}
\usepackage{colortbl}
\definecolor{maroon}{cmyk}{0,0.87,0.68,0.32}

\usepackage[acceptedWithA]{tacl2021v1}
\usepackage{xspace,mfirstuc,tabulary}

\newif\iftaclinstructions
\taclinstructionsfalse 
\iftaclinstructions
\renewcommand{\confidential}{}
\renewcommand{\anonsubtext}{(No author info supplied here, for consistency with
TACL-submission anonymization requirements)}
\newcommand{\instr}
\fi

\iftaclpubformat 

\else

\fi

\usepackage{amsthm} %
\theoremstyle{plain}

\theoremstyle{definition}

\theoremstyle{remark}

\title{Entropic Control for Homeostatic Optimization: Pre-empting Catastrophic Forgetting via Dynamic Feedback}

\author{
  \textbf{Shangxi Wu}$^{1}$\quad
  \textbf{Xiaorong Yan}$^{2}$\quad
  \textbf{Xiaojun Meng}$^{3}$\quad
  \textbf{Yuxiang Zhang}$^{1}$\quad
  \textbf{Jiansheng Wei}$^{3}$
  \\[4pt]
  {\normalfont $^{1}$Beijing Jiaotong University\quad
  $^{2}$The Hong Kong Polytechnic University\quad
  $^{3}$Huawei Noah's Ark Lab}
  \\[2pt]
  {\normalfont\small \texttt{\{wushangxi,yuxiangzhang\}@bjtu.edu.cn},\;
  \texttt{yanxiaorong027@gmail.com},\;
  \texttt{\{mengxiaojun,weijiansheng\}@huawei.com}}
}

\date{}

\begin{document}
\maketitle
\thispagestyle{plain}
\begin{abstract}
Supervised Fine-Tuning (SFT) of LLMs suffers from catastrophic forgetting, and prevailing mitigation strategies struggle to reconcile the plasticity-stability dilemma. To address this, we propose Entropic Control for Homeostatic Optimization (ECHO), a closed-loop control framework. By decomposing the KL divergence, we identify output entropy as a dominant observable component of KL variation. This enables us to leverage entropy as a computationally efficient online proxy to monitor forgetting risks in real-time, bypassing expensive reference-model inference. Consequently, ECHO employs an elastic controller that dynamically modulates the gradient weight of replay data via a proportional control law, using deviations from the pre-trained entropic equilibrium as an error signal. Empirical evaluations across 0.6B to 9B parameter architectures demonstrate that ECHO lies on, and in several comparisons extends, the observed plasticity-stability frontier and maintains high operational robustness against varying data mixing ratios.
\end{abstract}

\section{Introduction}
\label{sec:intro}

Large language models (LLM) have acquired extensive linguistic knowledge, world knowledge, and certain reasoning capabilities through autoregressive pre-training on massive corpora~\cite{DBLP:conf/nips/BrownMRSKDNSSAA20,DBLP:journals/corr/abs-2001-08361,DBLP:journals/tmlr/WeiTBRZBYBZMCHVLDF22}. To make models better follow human instructions and adapt to specific downstream tasks, Supervised Fine-Tuning has become one of the most commonly used alignment paradigms~\cite{DBLP:conf/nips/Ouyang0JAWMZASR22,DBLP:conf/iclr/WeiBZGYLDDL22,DBLP:conf/acl/WangKMLSKH23}. However, SFT also brings a long-standing and acute problem: Catastrophic Forgetting (CF)~\cite{DBLP:conf/nips/French93,DBLP:journals/corr/abs-2308-08747}. Under continuous downstream gradient updates, the model often significantly loses the general capabilities acquired during the pre-training stage while improving performance on new tasks, thus forming the plasticity-stability dilemma~\cite{DBLP:journals/corr/KirkpatrickPRVD16,DBLP:journals/pami/LiH18a}.

Mainstream mitigation strategies primarily fall into two categories. The first involves regularization via distribution constraints, typically adding KL penalties relative to a reference model to limit output deviation~\cite{DBLP:journals/corr/SchulmanWDRK17,DBLP:conf/nips/Ouyang0JAWMZASR22,DBLP:conf/nips/RafailovSMMEF23}. While effective for stabilization, this approach incurs substantial computational and memory overhead by requiring a resident reference model, limiting its viability in full-parameter fine-tuning. The second category is data replay, which mixes pre-training corpora with task data to suppress drift~\cite{DBLP:conf/cvpr/RebuffiKSL17,DBLP:conf/emnlp/ScialomCM22}. Although implementation-friendly, existing methods rely on static mixing ratios that overlook training dynamics: early stages demand strong drift suppression, whereas persistent high replay in later stages hinders new task adaptation. Consequently, static mixing suboptimally reduces a time-varying control problem to a rigid hyperparameter selection~\cite{DBLP:conf/nips/Xie0DDLLLL0Y23,DBLP:conf/nips/ChenRBWZSR23,DBLP:conf/iclr/JainCWKCSBKSSGG24}.

To efficiently resolve this dilemma, we revisit the theoretical foundations of distribution constraints. Although KL divergence is widely regarded as the canonical metric for quantifying distribution shift, its calculation imposes a substantial computational burden by requiring the concurrent maintenance of a reference model~\cite{DBLP:journals/corr/SchulmanWDRK17}. Through a theoretical decomposition of the KL divergence, we isolate its two constituent components: the cross-entropy between the reference and current policies, and the negative entropy of the current policy itself. Our empirical analysis yields a critical observation that the cross-entropy term remains comparatively stable throughout the fine-tuning process. In contrast, the fluctuations in output entropy account for the larger share of KL variation. This finding indicates that output entropy functions as a lightweight, reference-free proxy for monitoring the representational integrity of the model~\cite{DBLP:conf/iclr/HoltzmanBDFC20}. Consequently, we reframe the fine-tuning process from a static optimization problem into a dynamic control system. The primary objective shifts to minimizing task-specific loss while simultaneously constraining the entropic state of the model within a bounded equilibrium region derived from pre-training. This effectively suppresses implicit distribution drift without incurring the computational overhead associated with external reference models~\cite{DBLP:conf/iclr/PereyraTCKH17}.

Guided by this formulation, we propose \textbf{Entropic Control for Homeostatic Optimization (ECHO)}, a dynamic framework that turns data replay into feedback control. Unlike static mixing, ECHO monitors the absolute deviation between current prediction entropy and the pre-trained baseline. This introduces an elastic restoring force into optimization: the controller relaxes constraints to prioritize plasticity when the model is stable, but amplifies replay weight to pull parameters back toward the pre-training distribution when abnormal entropy deviations arise. This converts manual ratio tuning into an automated, self-correcting loop. The main contributions are as follows:
\begin{itemize}
    \item We decompose the KL divergence and empirically show that output entropy accounts for the larger share of its variation, validating entropy as a lightweight, reference-free proxy for monitoring fine-tuning health.
    \item We introduce ECHO, a plug-and-play endogenous self-correction mechanism that treats the data replay weight as a control variable directly driven by entropy feedback.
    \item Extensive evaluations across diverse datasets show that ECHO consistently outperforms static mixing baselines, lying on, and in several comparisons extending, the observed plasticity-stability frontier alongside computation-heavy regularization methods, but with virtually negligible computational and memory overhead during optimization.
\end{itemize}

\section{Related Works}
\label{sec:related_work}

\subsection{Regularization \& Alignment}
\label{sec:rw_regularization}

Regularization for continual learning traditionally relies on weight-space penalties, such as EWC~\cite{DBLP:journals/corr/KirkpatrickPRVD16}, SI~\cite{DBLP:conf/icml/ZenkePG17}, and MAS~\cite{DBLP:conf/eccv/AljundiBERT18}, or function-space distillation methods like LwF~\cite{DBLP:journals/pami/LiH18a}. In LLM fine-tuning, bounding the output distribution via KL divergence against a pre-trained reference model is standard practice~\cite{DBLP:conf/nips/Ouyang0JAWMZASR22}, yet it effectively doubles computational and memory overhead. While recent reference-free alignment approaches, including ORPO~\cite{DBLP:conf/emnlp/HongLT24} and SimPO~\cite{DBLP:conf/nips/0001X024}, along with weight deviation penalties like RecAdam~\cite{DBLP:conf/emnlp/ChenHCCLY20}, mitigate these costs, they fundamentally rely on static coefficients or preset schedules. Consequently, they cannot regulate constraint intensity in response to real-time drift.

\subsection{Data Replay \& Mixing}
\label{sec:rw_replay_mixing}

Data replay is ubiquitous for mitigating forgetting, though standard industrial practices rely on static mixing ratios that rigidly constrain the plasticity-stability trade-off. Recent data-centric frameworks, notably DoReMi~\cite{DBLP:conf/nips/Xie0DDLLLL0Y23} and Data Mixing Laws~\cite{DBLP:conf/iclr/YeLSZZQ25}, formally establish the impact of data weighting, with empirical scaling laws demonstrating that injecting even minimal pre-training data significantly curbs representational drift~\cite{DBLP:conf/icml/BethuneGBGCA25}. Advancing beyond static ratios, methods such as SuRe~\cite{hazard2025sure} attempt to dynamically select replay samples based on surprise signals. Nevertheless, these strategies remain primarily open-loop. They crucially lack an intrinsic, closed-loop feedback mechanism capable of modulating replay intensity based on the model's instantaneous stability.

\section{Methodology}
\label{sec:method}
In this section, we introduce ECHO, which reframes the plasticity-stability dilemma as a closed-loop control problem. We first identify output entropy as a reference-free indicator of distribution drift (\S\ref{sec:method_principle}), and then design a feedback controller that modulates the replay weight according to entropy deviation (\S\ref{sec:method_framework}).

\subsection{Motivation and Empirical Validation}
\label{sec:method_principle}

\begin{figure}[t]
\centering
\includegraphics[width=.9\linewidth]{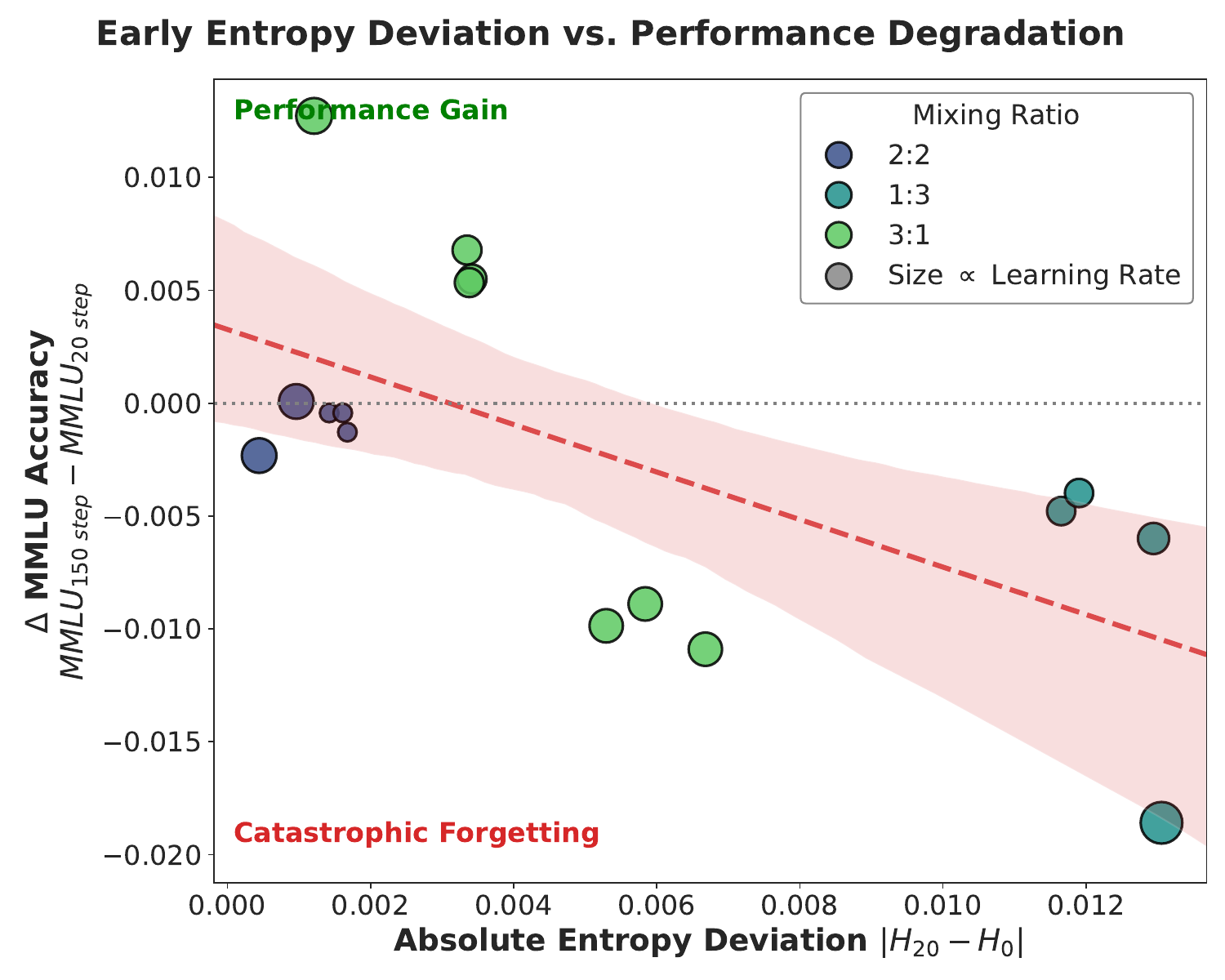}
\begin{minipage}{1.\linewidth}
    \centering
    \small
    \renewcommand{\arraystretch}{1.2}
    \setlength{\tabcolsep}{12pt}
    \begin{tabular}{l c}
        \toprule
        \textbf{Statistical Metric} & \textbf{Value} \\
        \midrule
        Pearson Correlation ($r$) & -0.617 \\
        Statistical Significance ($p$) & 0.014 \\
        \bottomrule
    \end{tabular}
\end{minipage}
\caption{Correlation analysis: early entropy deviation vs. final performance degradation. Results demonstrate that deviation from the entropy baseline $H_{\text{ref}}$ is strongly correlated with the loss of general capabilities.}
\label{fig:pilot_entropy_forgetting}
\end{figure}

Let the pre-trained reference model be denoted by $p_{\text{ref}}$ and the fine-tuned model at step $t$ by $p_{\theta_t}$. In standard Supervised Fine-Tuning, the objective is to optimize performance on a specific downstream dataset $\mathcal{D}_{\text{sft}}$. To mitigate the risk of catastrophic forgetting, a common practice involves constraining the model output on an anchor dataset $\mathcal{D}_{\text{anc}}$ to remain close to $p_{\text{ref}}$. The most rigorous measure for this distribution shift is the KL divergence averaged over the anchor distribution:
\begin{equation}
\label{eq:drift_def}
\mathcal{K}_t = \mathbb{E}_{x \sim \mathcal{D}_{\text{anc}}} \left[ D_{\mathrm{KL}}(p_{\theta_t}(\cdot|x) \,\|\, p_{\text{ref}}(\cdot|x)) \right].
\end{equation}
While theoretically sound, directly minimizing $\mathcal{K}_t$ as a regularizer incurs substantial computational overhead. It necessitates a synchronous forward pass through the frozen reference model for every training batch, which effectively doubles the memory footprint and significantly increases the wall-clock time per iteration. To address this efficiency bottleneck, we decompose the KL divergence into its constituent mathematical components:
\begin{equation}
\label{eq:kl_decomp}
\mathcal{K}_t = \underbrace{\mathbb{E}_{x \sim \mathcal{D}_{\text{anc}}} \left[ H_{\times}(p_{\theta_t}, p_{\text{ref}}) \right]}_{\text{Cross-Entropy Term}} - \underbrace{\mathbb{E}_{x \sim \mathcal{D}_{\text{anc}}} \left[ H(p_{\theta_t}) \right]}_{\text{Output Entropy } \mathcal{H}_t},
\end{equation}
where $H_{\times}$ denotes the cross-entropy and $H$ denotes the Shannon entropy. Equation~\ref{eq:kl_decomp} indicates that the total distribution drift $\mathcal{K}_t$ is jointly governed by two distinct factors: the alignment of the predictive mean with the reference distribution and the intrinsic uncertainty of the current policy.

\begin{figure*}[t]
\centering
\includegraphics[width=.9\linewidth]{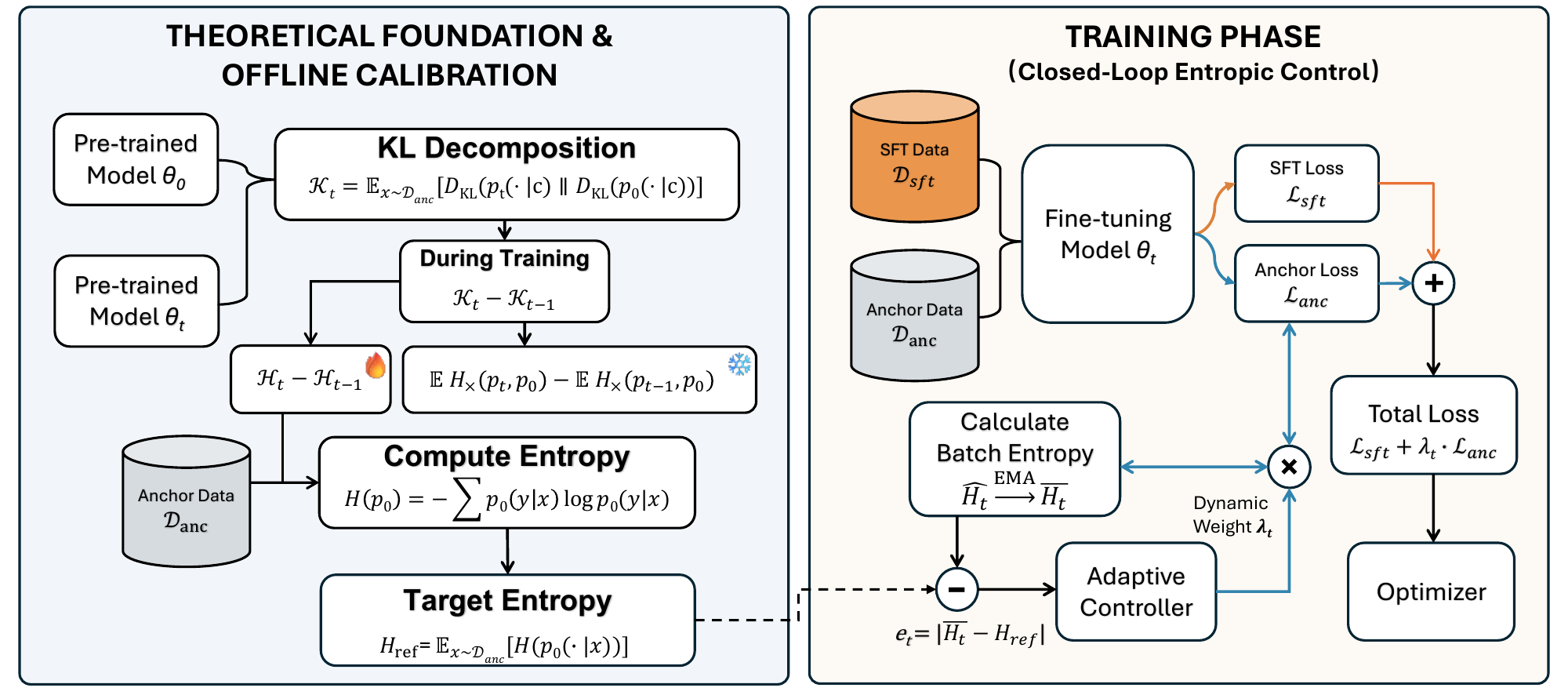}
\caption{Overview of the Entropic Control for Homeostatic Optimization.}
\label{framework}
\end{figure*}

\noindent\textbf{Theoretical Intuition.} The cross-entropy $H_{\times}$ measures alignment between the fine-tuned policy and reference likelihoods. Since SFT on high-quality instructions mainly adapts response style while preserving the syntax and world knowledge of the pre-trained backbone, this component is usually stable in the early stages.

In contrast, the self-entropy term $\mathcal{H}_t$ characterizes the spread of the distributional landscape. Catastrophic forgetting often induces two failure modes: diversity collapse, where the model overfits to narrow patterns and yields abnormally low entropy, and semantic dissolution, where internal representations drift toward high-entropy uniformity. Thus, $\mathcal{H}_t$ serves as a sensitive indicator of representational degradation. Since cross-entropy is more stable, regulating entropic volatility addresses the larger share of $\mathcal{K}_t$'s variation, motivating our concept of Entropic Homeostasis.

\noindent\textbf{Experimental Setup.} To verify this hypothesis and quantify the relative contribution of the two terms, we conducted a controlled study using Qwen3-0.6B. The protocol involved a grid search across 32 configurations, varying learning rates and mixing ratios between instruction-specific and anchor data, ensuring the observations are not tied to a specific hyperparameter setting. We monitored two variables: early entropy deviation, defined as $|H_{20} - H_{\text{ref}}|$, and general capability retention, measured by MMLU accuracy at step 150.

\noindent\textbf{Result Analysis.} As illustrated in Figure~\ref{fig:pilot_entropy_forgetting}, the empirical data reveals a moderate-to-strong and statistically significant negative correlation between the magnitude of early entropy deviation and the final performance retention. This robust inverse relationship suggests that detecting entropy drift serves as a reliable predictive proxy for anticipating catastrophic forgetting before the representational shift becomes irreversible. 

Furthermore, we quantified the specific contribution of each term in Equation~\ref{eq:kl_decomp} to the total distribution drift. By measuring the mean absolute deviation of both output entropy and cross-entropy relative to their initial values across all 32 experimental runs, we can compare the contributions of the two terms to KL variation.

\begin{table}[t]
\centering
\caption{\textbf{Distribution Shift Metrics.} We report entropy and cross-entropy deviations during training, showing that entropy varies more than cross-entropy.}
\label{tab:entropy_vs_ce_stats}
\resizebox{0.95\linewidth}{!}{
\begin{tabular}{l c c}
\toprule
\textbf{Metric} & \textbf{Mean $|\Delta|$ ($\pm$ Std)} & \textbf{Ratio vs. $H_{\times}$} \\
\midrule
Cross-Entropy ($H_{\times}$) & $0.0363 \pm 0.027$ & $1.00\times$ \\
\textbf{Output Entropy ($H$)} & $\mathbf{0.0559} \pm 0.024$ & $\mathbf{1.54\times}$ \\
\midrule
\multicolumn{3}{l}{\textit{Statistical Significance (Paired t-test):} $t=13.51$, $\bm{p < 10^{-13}}$ ***} \\
\bottomrule
\end{tabular}
}
\end{table}

The statistical analysis in Table~\ref{tab:entropy_vs_ce_stats} reveals a clear stability disparity. The mean absolute deviation of the output entropy term is approximately 1.54$\times$ that of the cross-entropy term, indicating that during fine-tuning, the shape of model uncertainty changes far more than its fundamental semantic alignment. A paired t-test yields a p-value below $10^{-13}$, confirming that entropy fluctuations are a dominant observable component of KL variation in the SFT settings studied.

This discovery offers a key computational advantage. While cross-entropy remains expensive because it requires a frozen reference model, output entropy $\mathcal{H}_t$ is directly produced during training from the current model logits already computed in the anchor-loss forward pass. Thus, constraining $\mathcal{H}_t$ near its pre-trained equilibrium provides a practical way to curb distribution drift without additional memory or computational overhead. Entropy matching is necessary but not sufficient for distributional closeness, so ECHO uses it as a feedback signal rather than a guarantee.

\subsection{The ECHO Framework}
\label{sec:method_framework}

Building on the observation that entropy deviation tracks distribution drift, we propose the ECHO framework. This system reformulates the fine-tuning process as a closed-loop feedback control problem where the replay intensity is modulated based on real-time state estimation. The overall procedure is detailed in Figure~\ref{framework}.

The observable variable in our control loop is the prediction entropy on the anchor distribution. For a batch of data $\mathcal{B}_{\text{anc}}$, we estimate the current entropic state $\widehat{H}_t$ as the expectation of the entropy over the valid tokens in the sequence:
\begin{equation}
\label{eq:entropy_est}
\widehat{H}_t = \mathbb{E}_{x \sim \mathcal{B}_{\text{anc}}} \left[ H(p_{\theta_t}(\cdot|x)) \right].
\end{equation}
This calculation is computationally efficient because it only requires the logits produced during the forward pass of the anchor loss. 

\noindent\textbf{Offline Reference Calibration.} Before fine-tuning, we establish a homeostatic setpoint denoted as $H_{\text{ref}}$. We perform a one-time forward pass using the pre-trained model $\theta_0$ on a subset of anchor data to obtain the target equilibrium:
\begin{equation}
H_{\text{ref}} = \mathbb{E}_{x \sim \mathcal{D}_{\text{anc}}} \left[ H(p_{\theta_0}(\cdot|x)) \right].
\end{equation}
This scalar value represents the intrinsic uncertainty level of the original model. It is frozen and serves as the definitive reference point for the subsequent controller throughout the training.

\subsubsection{The Elastic Feedback Loop}
During training, the system continuously monitors the deviation between the current state and the reference. We define the error signal $e_t$ as the absolute deviation to ensure that we penalize both distribution collapse and high-entropy uniformity.

\noindent\textbf{Elastic Actuation with Signal Smoothing.} To mitigate the inherent stochastic noise associated with mini-batch optimization, we refrain from feeding the instantaneous error signal directly into the controller. Rapid fluctuations in batch composition can lead to erratic adjustments of the replay weight. To prevent this, we implement an EMA to estimate the smoothed entropy state $\bar{H}_t$. This acts as a low-pass filter that isolates the persistent drift trend from high-frequency batch variance:
\begin{equation}
\label{eq:smoothing}
\bar{H}_t = \beta \cdot \bar{H}_{t-1} + (1 - \beta) \cdot \widehat{H}_t,
\end{equation}
where $\beta \in [0, 1)$ is the momentum coefficient that controls the smoothing horizon. The error signal is then computed as $e_t = | \bar{H}_t - H_{\text{ref}} |$.

Based on this filtered signal, the replay weight $\lambda_t$ follows an elastic stiffness law:
\begin{equation}
\label{eq:control_law}
\lambda_t = \mathcal{S}\left(\lambda_{\text{base}} + K_p \cdot e_t\right).
\end{equation}
Here, $\lambda_{\text{base}}$ serves as a minimal basal rate ensuring basic data exposure to the anchor set, and $K_p$ represents the elastic stiffness of the system. To ensure numerical stability and prevent the replay loss from overwhelming the gradients, the saturation operator $\mathcal{S}(z) = \mathrm{clamp}(z, \lambda_{\min}, \lambda_{\max})$ bounds $\lambda_t$ within $[\lambda_{\min}, \lambda_{\max}]$ (line~14 of Algorithm~\ref{alg:echo}). Note that $\lambda_{\text{base}}$, $K_p$, $\beta$, $\lambda_{\min}$, and $\lambda_{\max}$ are fixed hyperparameters, whereas $\bar{H}_t$, $e_t$, and $\lambda_t$ are state variables driven deterministically by the observed entropy; the controller has no learnable parameters. Crucially, this formulation reduces the complexity of hyperparameter tuning. While the optimal static mixing ratio often varies drastically across different datasets and learning rates, $K_p$ characterizes the sensitivity to distribution drift, and the effective weight $\lambda_t$ auto-adjusts with the magnitude of the smoothed error $e_t$. In practice, $K_p$ was set by a one-time sensitivity sweep on Qwen3-1.7B (Section~\ref{sec:exp_sensitivity_hyperparams}) and kept fixed in all experiments, without task- or backbone-specific retuning.

\subsubsection{Optimization Objective}
The objective function couples the anchor constraint to the current stability of the model by incorporating an elastic feedback-driven weight that responds to smoothed entropic deviations:
\begin{equation}
\mathcal{L}_{\text{total}} = \mathcal{L}_{\text{sft}}(\mathcal{B}_{\text{sft}}) + \lambda_t \cdot \mathcal{L}_{\text{anc}}(\mathcal{B}_{\text{anc}}).
\end{equation}
This mechanism creates a self-regulating optimization landscape. When the model remains stable and the error $e_t$ is near zero, $\lambda_t$ relaxes to its basal level. This maximizes plasticity for learning new tasks. Conversely, when significant drift occurs and $e_t$ increases, the restoring force $K_p \cdot e_t$ dominates. This increases the weight of the anchor loss and pulls the model back toward the safe manifold of the pre-trained distribution. This achieves a balance between plasticity and stability that static data mixing cannot replicate.

The complete training procedure of ECHO, including entropy calibration, EMA-based state estimation, and feedback-driven replay weighting, is summarized in Algorithm~\ref{alg:echo}.

\begin{algorithm}[t]
   \caption{Entropic Control for Homeostatic Optimization}
   \label{alg:echo}
\begin{algorithmic}[1]
   \STATE \textbf{Input:} Pre-trained model $\theta_0$, SFT dataset $\mathcal{D}_{\text{sft}}$, Anchor dataset $\mathcal{D}_{\text{anc}}$
   \STATE \textbf{Parameters:} Basal weight $\lambda_{\text{base}}$, Gain $K_p$, Bounds $[\lambda_{\min}, \lambda_{\max}]$, Momentum $\beta \in [0,1)$
   \STATE \textbf{Calibration:} Compute $H_{\text{ref}} \leftarrow \mathbb{E}_{x \sim \mathcal{D}_{\text{anc}}} [H(p_{\theta_0}(\cdot|x))]$
   \STATE Initialize $\theta \leftarrow \theta_0$, smoothed entropy $\overline{H}_0 \leftarrow H_{\text{ref}}$
   \FOR{step $t = 1$ {\bfseries to} $T_{\text{steps}}$}
       \STATE Sample batches $\mathcal{B}_{\text{sft}} \sim \mathcal{D}_{\text{sft}}$, $\mathcal{B}_{\text{anc}} \sim \mathcal{D}_{\text{anc}}$
       \STATE \textit{// Forward pass on anchor data}
       \STATE Compute logits and loss $\mathcal{L}_{\text{anc}}$ on $\mathcal{B}_{\text{anc}}$
       \STATE \textit{// State Estimation with EMA Smoothing}
       \STATE Estimate current batch entropy $\widehat{H}_t = \mathbb{E}_{x \sim \mathcal{B}_{\text{anc}}} [H(p_{\theta_t}(\cdot|x))]$
       \STATE Update smoothed state: $\overline{H}_t \leftarrow \beta \cdot \overline{H}_{t-1} + (1 - \beta) \cdot \widehat{H}_t$
       \STATE Compute error signal $e_t \leftarrow | \overline{H}_t - H_{\text{ref}} |$
       \STATE \textit{// Feedback Control \& Optimization}
       \STATE Compute weight $\lambda_t \leftarrow \text{clamp}(\lambda_{\text{base}} + K_p \cdot e_t, \lambda_{\min}, \lambda_{\max})$
       \STATE $\mathcal{L}_{\text{total}} \leftarrow \mathcal{L}_{\text{sft}} + \lambda_t \cdot \mathcal{L}_{\text{anc}}$
       \STATE Update $\theta \leftarrow \text{Optimizer}(\nabla \mathcal{L}_{\text{total}})$
   \ENDFOR
\end{algorithmic}
\end{algorithm}

\section{Experiments}
\label{sec:exp}

This section presents an empirical assessment of the proposed \textbf{ECHO} framework across high-interference fine-tuning scenarios. We evaluate three representative domains: code generation, mathematical, and logical inference, each requiring substantial parametric adaptation and thus prone to catastrophic forgetting. By reframing alignment as closed-loop homeostatic optimization, we address the following research questions:

\begin{itemize}[leftmargin=*]
    \item \textbf{RQ1: Disturbance Rejection \& Stability.} How does the entropic loop suppress drift and preserve general capabilities?
    \item \textbf{RQ2: Dynamic Tracking \& Plasticity.} Can the ECHO controller retain sufficient gradient plasticity to reach or exceed unconstrained fine-tuning on downstream objectives?
    \item \textbf{RQ3: Operational Robustness.} Does the entropy-driven mechanism remain effective across hyperparameters such as data mixing ratios, compared to static baselines or reference models?
\end{itemize}

\subsection{Experimental Setup}
\label{sec:exp_setup}

\paragraph{Models and Baselines.}
To evaluate the universality and scalability of our approach, we employed \textbf{Qwen3-0.6B}~\cite{DBLP:journals/corr/abs-2505-09388} and \textbf{Llama-3.2-3B}~\cite{DBLP:journals/corr/abs-2407-21783} as the primary foundation models. Additionally, we extended our analysis to \textbf{Qwen3-1.7B}, as well as larger-scale architectures including \textbf{Qwen3-8B} and \textbf{Gemma2-9B}~\cite{team2024gemma}, to verify the method's robustness and scaling potential across different parameter scales and tokenizer designs. We compared ECHO against five representative baselines:

\begin{itemize}[leftmargin=*]
    \item \textbf{Vanilla SFT:} Performs full-parameter fine-tuning on task-specific data. This serves as a control group to measure the upper bound of forgetting when stability constraints are absent.
    \item \textbf{Static Mixing~\cite{DBLP:conf/emnlp/ScialomCM22}:} Incorporates anchor data at a fixed replay ratio, representing industrial practice to mitigate forgetting via an open-loop and fixed replay schedule.
    \item \textbf{Data Mixing Laws:} Searches for the data composition ratio based on pilot experiments with a 0.6B model, fitting performance curves to balance task performance and capability retention.
    \item \textbf{LoRA~\cite{DBLP:conf/iclr/HuSWALWWC22}:} Freezes the backbone and optimizes low-rank adapters. By confining updates to a low-dimensional manifold, it serves as a resource-efficient baseline that implicitly mitigates drift through structural constraints.
    \item \textbf{KL-SFT:} Adds a KL divergence term toward the frozen base model. This bounds deviation from the pre-trained distribution to preserve capabilities.
\end{itemize}

\paragraph{Datasets.}
Our experiments spanned three independent domains, consistently using Alpaca-GPT4~\cite{DBLP:journals/corr/abs-2304-03277} as the anchor data source to preserve general instruction-following capabilities. The tasks were constructed as follows:
\begin{itemize}[leftmargin=*]
    \item \textbf{Code:} Uses a subset of Magicoder Evol Instruct~\cite{wei2023magicoder}, focusing on code logic and generation capabilities.
    \item \textbf{Math:} Utilizes the Meta Math QA dataset~\cite{yu2023metamath}, containing complex mathematical reasoning chains with step-by-step solutions.
    \item \textbf{Reasoning:} Employs the Open Platypus dataset~\cite{lee2023platypus}, focusing on logical inference and scientific reasoning.
\end{itemize}

\begin{table*}[t]
\centering
\small
\setlength{\tabcolsep}{4.5pt}
\caption{Main Results on Qwen3-0.6B and Llama-3.2-3B. We report raw scores and the improvement ($\Delta$) over the Base Model. Code, Math, and Reasoning are independent single-run experiments from the same checkpoint, each mixed with Alpaca-GPT4 anchor data; MMLU is averaged over the three. Bold: best; \underline{underline}: second best.}
\label{tab:main_results}
\begin{tabular}{@{}l rr rr rr rr@{}}
\toprule
\multirow{2}{*}{\textbf{Method}}
 & \multicolumn{2}{c}{\textbf{Code}}
 & \multicolumn{2}{c}{\textbf{Math}}
 & \multicolumn{2}{c}{\textbf{Reasoning}}
 & \multicolumn{2}{c}{\textbf{General Stability}} \\
\cmidrule(lr){2-3} \cmidrule(lr){4-5} \cmidrule(lr){6-7} \cmidrule(lr){8-9}
 & HumanEval\,$\uparrow$ & $\Delta$ & GSM8K\,$\uparrow$ & $\Delta$
 & HellaSwag\,$\uparrow$ & $\Delta$ & MMLU\,$\uparrow$ & $\Delta$ \\
\midrule
\multicolumn{9}{@{}l}{\emph{Backbone: Qwen3-0.6B}} \\
Base Model          & 18.90 & ---    & 10.39          & ---    & 47.35          & ---   & 49.17          & ---     \\
Vanilla SFT         & 31.10 & +12.20 & \underline{55.49} & +45.10 & 48.55       & +1.20 & 42.91          & $-$6.26 \\
Data Mixing Laws    & \underline{37.80} & +18.90 & 55.27 & +44.88 & \underline{50.81} & +3.46 & 43.38 & $-$5.79 \\
LoRA                & 23.17 & +4.27  & 17.51          & +7.12  & 48.79          & +1.44 & \textbf{44.81} & $-$4.36 \\
KL-SFT              & 35.61 & +16.71 & 53.35          & +42.96 & 48.09          & +0.74 & \underline{44.04} & $-$5.13 \\
\textbf{ECHO (Ours)}& \textbf{38.41} & +19.51 & \textbf{56.48} & +46.09 & \textbf{51.00} & +3.65 & 43.88 & $-$5.29 \\
\midrule
\multicolumn{9}{@{}l}{\emph{Backbone: Llama-3.2-3B}} \\
Base Model          & 26.83 & ---    & 9.78           & ---    & 55.88          & ---    & 55.22          & ---     \\
Vanilla SFT         & \underline{45.12} & +18.29 & \underline{58.83} & +49.05 & \underline{76.09} & +20.21 & 48.89 & $-$6.33 \\
Data Mixing Laws    & 44.51 & +17.68 & 55.57          & +45.79 & 75.99          & +20.11 & 51.04          & $-$4.18 \\
LoRA                & 31.70 & +4.87  & 45.64          & +35.86 & 73.97          & +18.09 & \textbf{55.04} & $-$0.18 \\
KL-SFT              & 44.63 & +17.80 & 55.29          & +45.51 & 71.05          & +15.17 & \underline{54.47} & $-$0.75 \\
\textbf{ECHO (Ours)}& \textbf{45.73} & +18.90 & \textbf{59.34} & +49.56 & \textbf{76.44} & +20.56 & 51.53 & $-$3.69 \\
\bottomrule
\end{tabular}
\end{table*}

\paragraph{Implementation Details.}
All models were trained by AdamW~\cite{DBLP:conf/iclr/LoshchilovH19} optimizer with a global batch size of 32. The learning rates were set to $5\times 10^{-5}$ for the 0.6B model and $3\times 10^{-5}$ for the 1.7B and 3B models. For ECHO, we set the control gain $K_p=5.0$, the base weight $\lambda_{\text{base}}=0.5$, the saturation limit $\lambda_{\max}=10.0$, and $\beta=0.9$ (see Section~\ref{sec:exp_sensitivity_hyperparams} for the sensitivity analysis). We employed the \texttt{lm-evaluation-harness} framework for assessment, using MMLU~\cite{DBLP:conf/iclr/HendrycksBBZMSS21} to gauge general capabilities, while HumanEval~\cite{DBLP:journals/corr/abs-2107-03374}, GSM8K~\cite{DBLP:journals/corr/abs-2110-14168}, and HellaSwag~\cite{DBLP:conf/acl/ZellersHBFC19} were used to evaluate domain-specific performance. For all mixed training setups, we constructed datasets totaling 40,000 samples to ensure consistent training steps across all methods.

\subsection{Main Results}
\label{sec:exp_main_results}

Table \ref{tab:main_results} presents the evaluation results across the Code, Math, and Reasoning domains. We analyze the performance from two complementary perspectives: plasticity and stability.

\paragraph{Plasticity: Adaptation to Downstream Tasks.}
A primary concern with regularization-based methods is whether constraints impede the learning of new knowledge. As shown in Table \ref{tab:main_results}, ECHO consistently achieves performance comparable to, and in several cases surpassing, unconstrained Vanilla SFT.
For instance, on the Llama-3.2-3B architecture, our method achieves the highest accuracy on the GSM8K Math task (59.34\%), outperforming Vanilla SFT (58.83\%). In contrast, Data Mixing Laws exhibits signs of under-fitting on this complex reasoning task (55.57\%), suggesting that a fixed replay ratio may introduce excessive gradient interference during critical learning phases.
Similarly, on the Qwen3-0.6B Code task, our method achieves a HumanEval pass@1 of 38.41\%, outperforming Vanilla SFT (31.10\%). This indicates that the dynamic weight permits the necessary parameter updates for domain adaptation without the rigid constraints of methods like LoRA, which lags in downstream performance.

\paragraph{Stability: Mitigation of Forgetting.}
Vanilla SFT suffers from catastrophic forgetting, evidenced by MMLU declines. While Data Mixing Laws and KL-SFT alleviate this, they often compromise downstream adaptation. On Llama-3.2-3B, although KL-SFT achieves the strongest general retention, it fails to match ECHO's specialized gains.
ECHO provides a more balanced trade-off by intensifying constraints only upon detecting entropy deviation. While LoRA maintains peak stability via its frozen backbone, its inability to master complex new tasks underscores the limits of structural constraints. ECHO lies on, and in several comparisons extends, the observed plasticity-stability frontier---maximizing task acquisition while keeping general capabilities within a robust homeostatic range.

\paragraph{Computational Cost.}
Unlike KL-SFT, ECHO needs neither a resident reference model nor an extra forward pass: the entropy comes from logits already computed for the anchor loss (Eq.~\ref{eq:entropy_est}), adding only an entropy reduction, an EMA update, and a scalar control law. In our runs, its memory footprint and wall-clock time were nearly identical to those of Static Mixing.

\subsection{Robustness to Data Mixing Ratios}
\label{sec:exp_sensitivity}

\begin{figure*}[!t]
\centering
\includegraphics[width=0.95\linewidth]{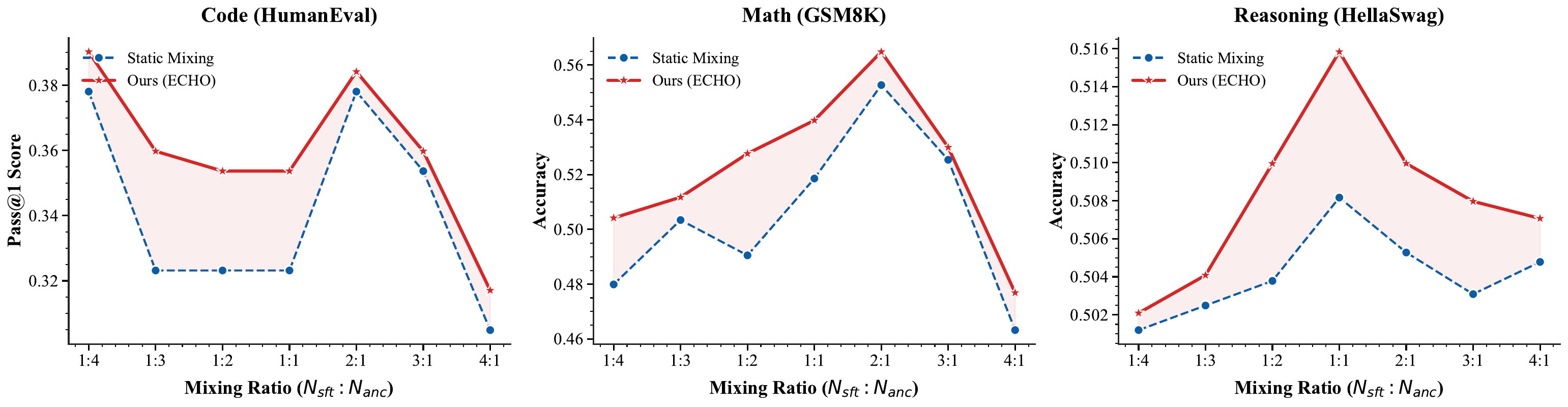}
\caption{Sensitivity Analysis of Data Proportioning. Performance across Code, Math, and Reasoning domains as mixing ratio $\gamma$ varies. ECHO matches or outperforms Static Mixing at every ratio in all three domains.}
\label{fig:ratio_sensitivity}
\end{figure*}

In replay-based fine-tuning, the mixing ratio of new and old data (Mixing Ratio, $\gamma = N_{\text{sft}} / N_{\text{anc}}$) often determines the model's trade-off between ``catastrophic forgetting'' and ``under-fitting''. To verify whether ECHO reduces dependence on this sensitive hyperparameter, we compared the performance boundaries of dynamic strategy and static mixing on Code, Math, and Reason domains by traversing $\gamma$ from $0.25$ to $4.0$.

Figure \ref{fig:ratio_sensitivity} shows that ECHO achieved downstream task performance superior to or equal to Static Mixing at every mixing ratio, consistently improving over the static baseline. In the Math reasoning task, ours improved GSM8K accuracy to $56.5\%$ at $\gamma=2$, higher than the baseline's $55.3\%$ at the same ratio, and maintained this advantage under high data sparsity at $\gamma=4$. This indicates that the dynamic weight introduced by ECHO is not a simple constraint, but optimizes the training trajectory through precise temporal adjustment, achieving better convergence without sacrificing new knowledge learning.

Both methods share a similar dependence on $\gamma$ (e.g., Math peaks at $\gamma=2$ and drops at $\gamma=4$), but ECHO is never worse than Static Mixing across all 21 configurations: rather than relying on a single ``sweet spot'', entropy feedback yields a consistent upward shift. This robustness stems from entropy feedback: when anchor data is scarce (large $\gamma$) and causes entropy drift, the controller raises $\lambda_t$ to strengthen protection. Even at $\gamma=4.0$, where replay data is severely lacking, ECHO retains its advantage on all three downstream tasks. Thus, whichever ratio is chosen, adding entropy feedback is a safe improvement over static mixing.

\subsection{Generalization and Scaling Analysis}
\label{sec:exp_scaling}

\begin{table}[t]
\centering
\small
\caption{Performance across Large-Scale Models. We report task-specific accuracy and MMLU scores. Results show that ECHO effectively scales to 8B and 9B parameters across different architectures. All results are from a single run.}
\label{tab:scaling_results}
\setlength{\tabcolsep}{4.5pt} 
\resizebox{\linewidth}{!}{
\begin{tabular}{ll cccc}
\toprule
\textbf{Backbone} & \textbf{Method} & \textbf{Code} & \textbf{Math} & \textbf{Reason} & \textbf{MMLU} \\
\midrule
\multirow{2}{*}{Qwen3-8B} & Static Mix & 64.02 & 80.89 & 78.74 & 61.61 \\
 & \textbf{ECHO (Ours)} & \textbf{68.90} & \textbf{81.80} & \textbf{78.88} & \textbf{74.55} \\
\cmidrule(lr){1-6}
\multirow{2}{*}{Gemma2-9B} & Static Mix & 51.83 & 77.79 & 79.83 & 69.47 \\
 & \textbf{ECHO (Ours)} & \textbf{53.66} & \textbf{78.54} & \textbf{80.63} & \textbf{70.00} \\
\bottomrule
\end{tabular}
}
\end{table}

To evaluate ECHO in complex multi-task environments, we conducted experiments using a balanced 1:1:1:1 mixture of coding, mathematics, reasoning, and a general anchor set. This setting simulates high-interference training, where models must acquire specialized skills without eroding foundational knowledge. The controller modulates the weight of the anchor stream relative to the three task-specific streams, and we compare against Static Mixing on the same mixture.

Results in Table~\ref{tab:scaling_results} show that ECHO outperforms static mixing on both the 8B and 9B backbones. In the 1:1:1:1 regime, static mixing struggles to balance competing domain requirements, often causing collapse in general knowledge, especially when rigid replay ratios fail against intensive downstream gradients. In contrast, ECHO identifies representational drift at each step and adjusts replay weights, shielding the pre-trained distribution while facilitating specialized skill acquisition. Results on Qwen3-8B and Gemma2-9B confirm cross-architecture robustness: despite differences in tokenizer designs and pre-training objectives, entropic signals remain informative, enabling feedback control without architecture-specific tuning. This highlights ECHO's scaling potential: by replacing manual data proportioning with closed-loop control, a single configuration can adapt to backbones while remaining on the observed plasticity-stability frontier.

\subsection{Qualitative Analysis}
\label{sec:exp_qualitative}

\begin{figure}[t]
\centering
\includegraphics[width=.85\linewidth]{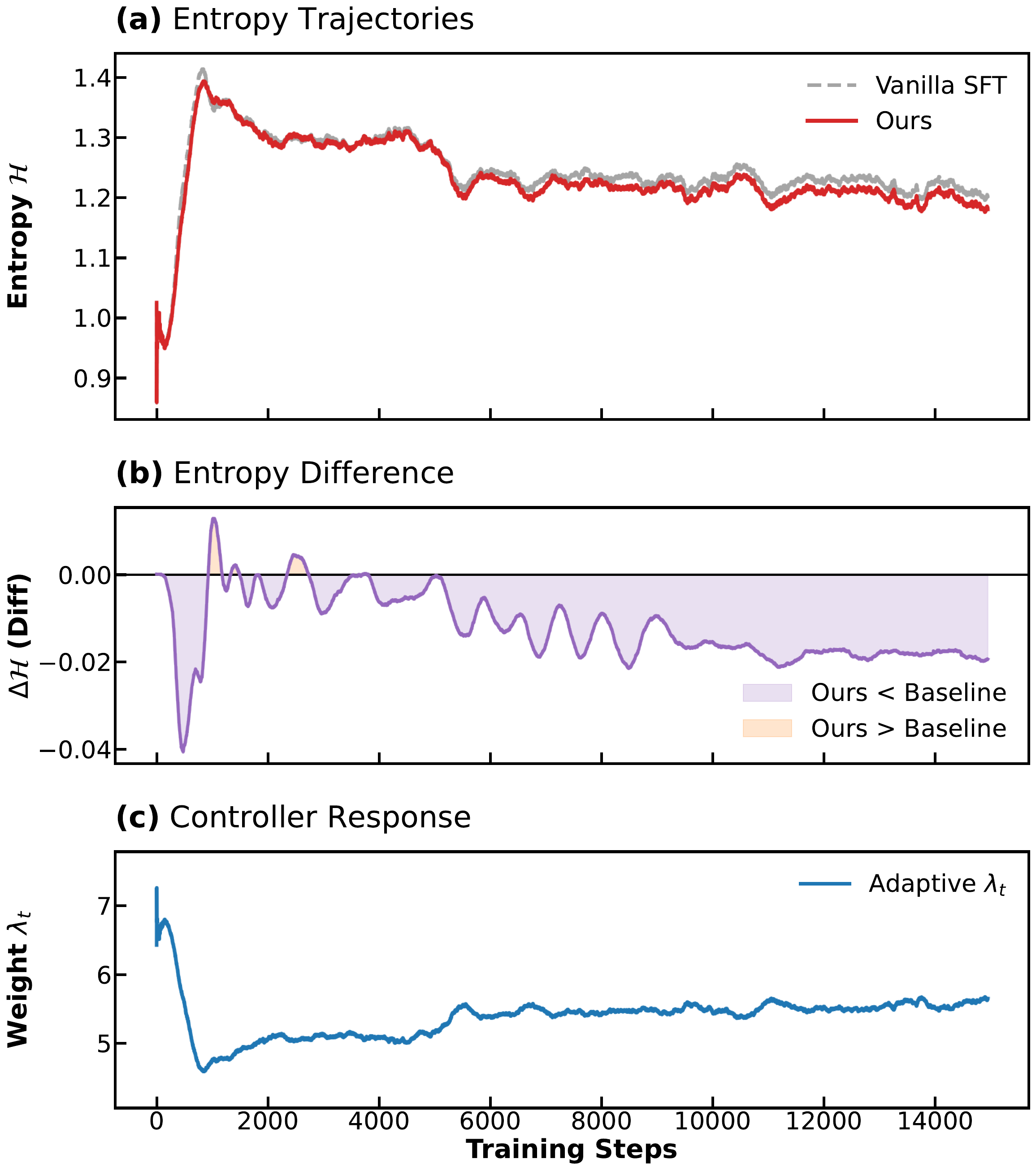}
\caption{Visualization of training dynamics. (a) Entropy trajectories: Ours maintains lower anchor-data uncertainty than Vanilla SFT. (b) Negative entropy difference indicates suppression of forgetting. (c) Adaptive weight $\lambda_t$ starts high to protect early stability, then adjusts to balance task learning and retention.}
\label{fig:dynamics}
\end{figure}

To understand the working mechanism of ECHO, we analyze the temporal dynamics of key indicators in training, as visualized in Figure \ref{fig:dynamics}.

\paragraph{Suppression of Entropy Drift.}
Figures \ref{fig:dynamics}(a) and (b) compare the entropy trajectories of our method against Vanilla SFT. In the uncontrolled baseline, the prediction entropy on anchor data rises, indicating that the model becomes increasingly uncertain about the pre-training distribution as it overfits to the new task. In contrast, our method maintains the entropy at a consistently lower level. The persistent negative difference in Figure \ref{fig:dynamics}(b) demonstrates that the feedback loop actively suppresses this uncertainty. By forcing the model to retain high confidence on anchor data, the system effectively locks the representational manifold in a safe homeostatic interval throughout the fine-tuning.

\paragraph{Adaptive Regulation Mechanism.}
The real-time response of the penalty weight $\lambda_t$, shown in Figure \ref{fig:dynamics}(c), reveals an intuitive progressive adaptation strategy. At the onset of training, we observe a sharp spike in $\lambda_t$. This high initial weight acts as a strong protective barrier, preventing the sudden distributional shock of new data from disrupting existing knowledge. As the model gradually reconciles the new instructions with its prior knowledge, $\lambda_t$ automatically stabilizes at a lower maintenance level. This transition allows the focus to shift progressively toward learning new downstream tasks while keeping anchor loss within a controlled bound. Crucially, this mechanism differs from simple learning rate decay; if the model encounters difficult batches that threaten stability later in training, the controller instantly increases $\lambda_t$ to re-impose constraints, ensuring long-term robustness without manual scheduling.

\section{Hyperparameter Sensitivity Analysis}
\label{sec:exp_sensitivity_hyperparams}

\begin{figure}[t]
\centering
\includegraphics[width=1.0\linewidth]{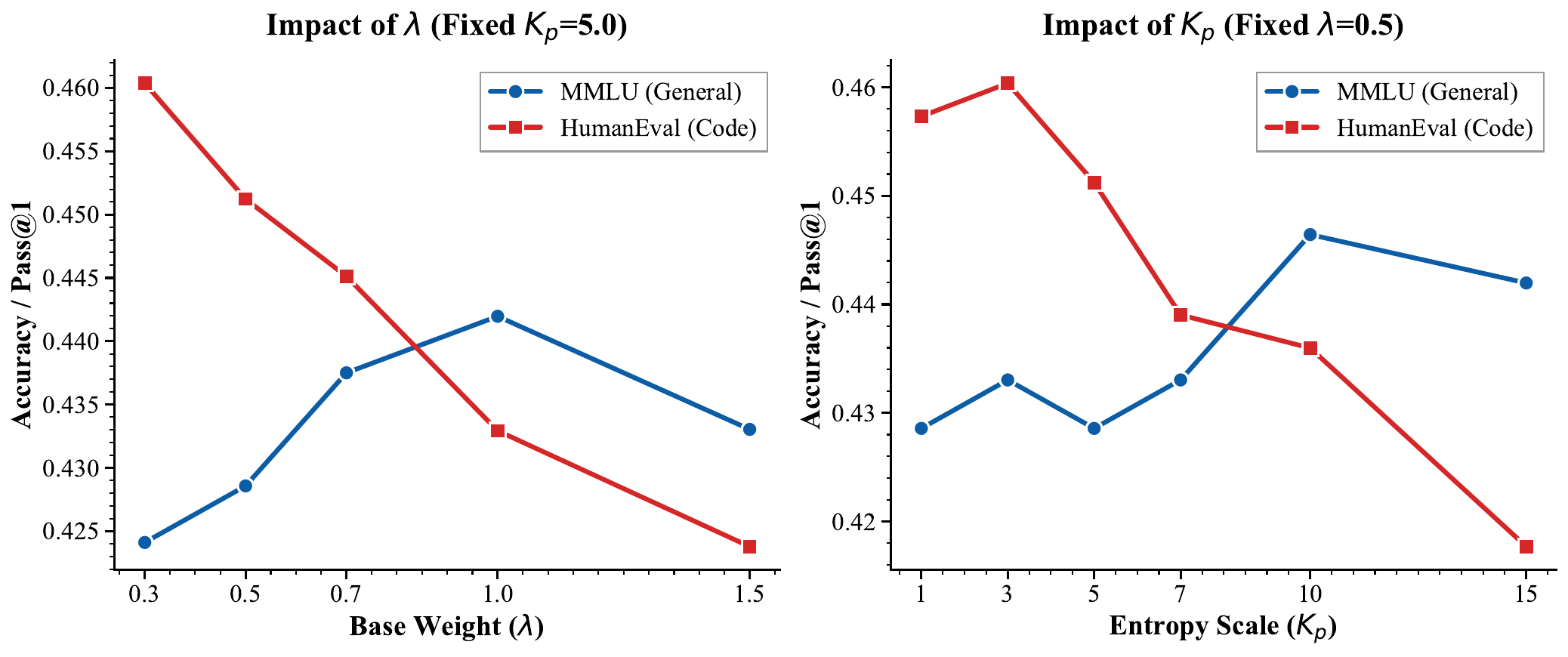}
\caption{Parameter sensitivity analysis on Qwen3-1.7B. Left: Base Weight $\lambda$ modulates the static stability-plasticity trade-off. Right: Entropy Scale $K_p$ acts as a dynamic regularizer; increasing $K_p$ restores general capabilities but may constrain new learning if the penalty becomes excessive.}
\label{fig:ablation}
\end{figure}

To further examine the robustness of ECHO, we conduct a hyperparameter sensitivity study on \textbf{Qwen3-1.7B}. We vary the basal anchoring weight $\lambda$ and the proportional gain $K_p$ to analyze how static anchoring and dynamic entropy feedback jointly regulate the stability-plasticity trade-off. Results are shown in Figure~\ref{fig:ablation}.

As illustrated in Figure \ref{fig:ablation} (Left), the Base Weight $\lambda$ governs the \textit{stability-plasticity dilemma}: increasing $\lambda$ from 0.3 to 1.0 raises general capability retention (MMLU, 42.4$\rightarrow$44.2), while new task acquisition (HumanEval) decreases monotonically from 46.0 to 42.4 over $\lambda \in [0.3, 1.5]$; beyond $\lambda=1.0$, MMLU no longer improves (43.3 at $\lambda=1.5$). This confirms that $\lambda$ explicitly controls the intransigence of the model against the pre-training distribution; a lower value facilitates rapid adaptation at the risk of forgetting, whereas a higher value enforces rigidity.

In contrast, the Entropy Scale $K_p$ exhibits a threshold-based regulatory behavior, acting as a dynamic safety valve (Figure \ref{fig:ablation}, Right). At lower scales, the feedback signal proves insufficient to counteract distribution drift, resulting in sub-optimal retention. Elevating $K_p$ to 10.0 imposes stricter entropic constraints and raises MMLU from 42.9 (at $K_p=5$) to 44.6 by penalizing deviations from the baseline manifold. However, excessive regularization narrows the optimization landscape, marginally hindering the learning of complex reasoning patterns required for code generation. Consequently, we identified the configuration of \textbf{$\lambda=0.5$} and \textbf{$K_p=5.0$} as default. This setting effectively mitigates catastrophic forgetting through sufficient entropic feedback without imposing excessive rigidity that would impede the acquisition of downstream competencies, thereby achieving a balanced trade-off between maintaining general knowledge and learning specific tasks.

\section{Conclusion}
\label{sec:conclusion}
In this paper, we address catastrophic forgetting in supervised fine-tuning with ECHO, a closed-loop framework that regulates replay weights via stability signals. By linking prediction entropy to distribution drift, ECHO uses output entropy as an efficient, reference-free KL surrogate, strengthening replay only under deviations from the pre-trained equilibrium. Experiments from \textbf{Qwen3-0.6B} to \textbf{Gemma2-9B} across math, code, and reasoning show that ECHO alleviates the ``plasticity-stability'' dilemma of static mixing, preserving general capabilities while improving downstream adaptation without frozen reference models.

\section*{Limitations}
This paper focuses on mitigating catastrophic forgetting during supervised fine-tuning and validates the effectiveness of ECHO across different model scales and multiple high-interference tasks. Nevertheless, several directions remain worth further exploration. First, we use output entropy as a lightweight stability signal to avoid the overhead of frozen reference models. While this signal is supported by our theoretical analysis and empirical results, future work can further examine its applicability under longer training horizons, larger model scales, and more complex data distributions. Second, our experiments mainly cover representative high-interference scenarios, including code generation, mathematical reasoning, and logical inference. Future work can extend ECHO to broader continual alignment settings, such as multilingual instruction tuning, safety preference optimization, and multimodal model updating. In addition, the current implementation adopts a simple proportional control law to regulate replay weights, which helps preserve efficiency and interpretability. Exploring adaptive control gains or richer control strategies may bring further gains in specific scenarios.

\section*{Impact Statement}
ECHO aims to make continual fine-tuning of large language models more stable and resource-efficient. By using an entropy-based feedback signal instead of a frozen reference model, ECHO reduces the memory and computational overhead required to preserve general capabilities during task adaptation. This can help researchers and practitioners with limited hardware resources update models more conveniently and provides a more sustainable solution for continual model optimization in specialized domains. The framework also supports more reliable deployment by mitigating catastrophic forgetting when models adapt to new tasks, thereby reducing the risk of unintended degradation in general capabilities after fine-tuning. Preserving general capabilities while improving downstream performance is practically valuable for long-term model maintenance and continual alignment. More broadly, ECHO offers a lightweight closed-loop perspective that encourages future research on adaptive training mechanisms balancing plasticity, stability, and efficiency.

\bibliography{tacl2021}
\bibliographystyle{acl_natbib}

\end{document}